\documentclass[sigconf,nonacm]{acmart}

\usepackage{amsmath, amssymb, graphicx,bm}
\usepackage{bbm}
\usepackage{mathtools}
\usepackage{algorithm}
\usepackage[noend]{algpseudocode}

\makeatletter
\def\BState{\State\hskip-\ALG@thistlm}
\makeatother

\DeclarePairedDelimiter\ceil{\lceil}{\rceil}
\DeclarePairedDelimiter\floor{\lfloor}{\rfloor}

\usepackage{amsfonts}       

\AtBeginDocument{%
  \providecommand\BibTeX{{%
    \normalfont B\kern-0.5em{\scshape i\kern-0.25em b}\kern-0.8em\TeX}}}

\begin{document}

\title{Factor Analysis of Mixed Data for Anomaly Detection}

\author{Matthew Davidow}
\email{mbd83@cornell.edu}
\affiliation{%
  \institution{Cornell University}
  \city{Ithaca}
  \state{New York}
  \postcode{14853}
}

\author{David S.\ Matteson}
\email{Matteson@cornell.edu}
\affiliation{%
  \institution{Cornell University}
  \city{Ithaca}
  \state{New York}
  \postcode{14853}
}

\renewcommand{\shortauthors}{Davidow and Matteson.}
\newif\ifcc
\newif\ifkeyword
\ccfalse
\keywordfalse

\begin{abstract}
Anomaly detection aims to identify observations that deviate from the typical pattern of data. 
Anomalous observations may correspond to financial fraud, health risks, or incorrectly measured data in practice. We show detecting anomalies in high-dimensional mixed data is enhanced through first embedding the data then assessing an anomaly scoring scheme. We focus on unsupervised detection and the continuous and categorical (mixed) variable case. 
We propose a kurtosis-weighted {\it Factor Analysis of Mixed Data}  for anomaly detection, {\it FAMDAD}, to obtain a continuous embedding for anomaly scoring. We illustrate that anomalies are highly separable in the first and last few ordered dimensions of this space, and test various anomaly scoring experiments within this subspace.  
Results are illustrated for both simulated and real datasets, and the proposed approach (FAMDAD) is highly accurate for high-dimensional mixed data throughout these diverse scenarios.  



\end{abstract}


\ifcc

\fi

\ifkeyword
\fi


\maketitle

\section{Introduction}
 Many anomaly detection tasks require analysis of high-dimensional mixed-type data; however, standard algorithms are unusable or require {\it ad hoc} workarounds for joint analysis of continuous and categorical variables. For example a common scheme is to dummy (one-hot) encode all categorical variables then employ algorithms for continuous data on the original continuous variables combined with the new (one-hotted) binary variables. However, this potentially leads to a huge dimension increase when there are copious categories, and the variable-wise embedding ignores the dependence structure among the original variables. 

We propose a kurtosis-weighted {\it Factor Analysis of Mixed Data} \cite{pages2014multiple} for anomaly detection embedding, \textbf{\emph{FAMDAD}}, in an unsupervised high-dimensional {\it mixed} (continuous and categorical) data setting.
FAMDAD builds a continuous, reduced dimension embedding that is amenably paired with popular detection algorithms for a low-dimensional continuous domain, including Isolation Forest (ISO) \cite{liu2008isolation,guha2016robust}, and Simple Probabilistic Anomaly Detector (SPAD) \cite{aryal2016revisiting}. While there are other popular unsupervised anomaly detection algorithms such as Local Outlier Factor (LOF) \cite{breunig1999optics}, and DBSCAN \cite{ester1996density}, we focus on the embedding property of FAMDAD and pair it with the fast, simple scoring algorithms of ISO and SPAD. Isolation Forest randomly partitions the data, recursing until each datapoint is in its own partition. Anomalies are likely to become isolated in their own partition earlier in this process, since they do not have neighbors who are similar to them. Simple probability anomaly detector scores observations by first discretizing any continuous columns, then gives each coordinate a score based on its frequency. The final score for each datapoint is the product of the scores along each dimension, which is similar to the naive Bayes independence assumption.
%
%
 %

Anomaly detection has seen an increased interest in recent years with widespread use in financial fraud detection \cite{ahmed2016survey}, medical applications \cite{salem2013anomaly}, server attack monitoring \cite{lakhina2004diagnosing}, and its general application as a pre-processing step to clean datasets  \cite{bretas1989iterative}. 
There are no widely agreed upon formal definitions of anomalies as the concept proves too elusive and contextually dependent, 
but an informal definition is often cited by Hawkins (1980): ``An outlier is an observation that deviates so much from other observations as to arouse suspicious that it was generated by a different mechanism." 

Supervised detection methods are often not applicable due to the difficulty of obtaining anomaly labeled data for training \cite{aggarwal2015outlier}. Even when labeled data exists, highly imbalanced classes (single anomalies may have their own class) can lead to poor performance when supervised methods are implemented naively. Further, new anomalies may not conform to the patterns of training data anomalies, especially in adversarial settings such as fraud and server attacks \cite{lakhina2004diagnosing,ahmed2016survey}. Thus unsupervised anomaly detection methods have great practical importance. 
 In high dimensions, distance-based anomaly detection methods that are agnostic to the geometry of the space or heterogeneity of the data break down \cite{houle2013dimensionality}. 
 Parametric methods focus, for example, on robustly estimating a covariance matrix \cite{filzmoser2008outlier,xu2010robust,mccoy2011two,candes2011robust}, then scoring anomalies under a Gaussian distributional assumptions. Unsupervised clustering and mixture distribution methods are also widely used in practice, but these primarily appeal to the continuous variable setting where a Gaussian assumption is valid.

 Anomaly detection is more challenging when some anomalies are relatively near inliers, a problem known as {\it masking}. Additionally, anomalies can appear in clusters, with large clusters of anomalies not unlike an inlier cluster, a pattern known as {\it swamping} \cite{aggarwal2015outlier, chandola2009anomaly}. Anomaly detection is also made difficult by the imbalance of classes, and the vague definition of an anomaly. Mixed data presents an additional difficulty, as methods based on Euclidian and similar distances are haphazard.

 To overcome these challenges we propose a three step FAMDAD approach. First, we provide a kurtosis-weighted FAMD algorithm to embed high dimensional mixed data into a continuous space. Next, we extract a lower dimensional sub-space from this embedding on which anomalies are expected to be well separated. Finally, we employ two benchmark unsupervised anomaly detection algorithms for anomaly scoring on this anomaly embedding continuous sub-space: Simple Probabilistic Anomaly Detector (SPAD) and Isolation Forest (ISO).
 %
 %
 
\begin{figure}
\centering
\includegraphics[width=0.5\textwidth]{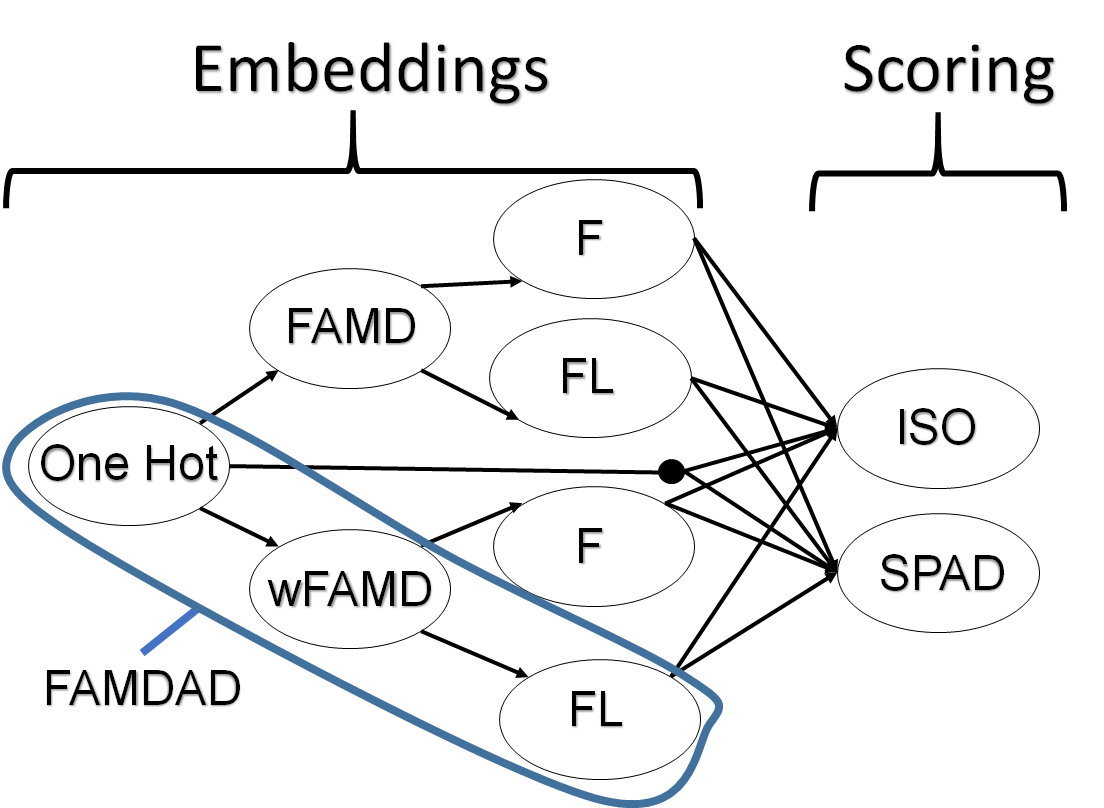}
\includegraphics[width=0.5\textwidth]{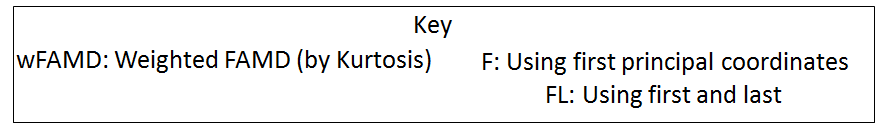}
\caption{\label{fig:Flow} Flow diagram showing comparison methods with embedding approaches on the left and scoring approaches on the right. The proposed FAMDAD algorithm utilizes the embedding depicted in the lowest track, one-hot $\rightarrow$ wFAMD $\rightarrow$ FL, paired with any scoring approach.}
\end{figure}

 We contrast each of these anomaly scoring approaches through the different stages of our anomaly embedding. Specifically, they are applied to (1) the mixed data directly (after one-hot transformation), (2) the data after applying the standard FAMD embedding, (3) the data embedded via a kurtosis-based weighted FAMD, and for the (4) FAMD and (5) kurtosis-weighted FAMD embeddings using an alternative sub-space selection strategy, where (5) defines the proposed FAMDAD embedding; see figure \ref{fig:Flow}.
 
 We show that FAMDAD has the power to embed anomalies in a reduced continuous sub-space that harnesses the joint associations across categorical and continuous variables, such that even a simple dimension-wise based algorithm such as SPAD performs well in this sub-space. 
%
Our main contributions include:
\begin{enumerate}
    \item{New unsupervised anomaly detection method for high-dimensional mixed data (continuous/categorical);}
    \item{Novel anomaly embedding algorithm via kurtosis-weighted factor model to isolate anomalies;}
    \item {Dimension reduction and sub-space selection approach for improved accuracy and efficiency;}
    \item{Comprehensive comparison of various anomaly scoring algorithms applied to  mixed data.}
\end{enumerate}

\section{Methodology}
The proposed FAMDAD method utilizes a factor analysis of mixed data \cite{pages2014multiple} approach as both a data transformation tool that faithfully embeds mixed data into a purely continuous representation, and as a dimension reduction tool. Kernel PCA and spectral methods such as diffusion maps are alternative approaches to transform data \cite{von2007tutorial, coifman2006diffusion}; however, we found a robust  FAMD approach proves superior in performance when combined with subspace selection, as it learns the main directions of variation among the embedded data. The proposed FAMDAD algorithm is detailed below.

We first present preliminary notation. We use boldface for vectors, capital letters for matrixes, and lowercase for scalars. Let $\mathbf{1}_n$ denote a length $n$ vector of ones, let $I_n$ denote the $n \times n$ identity matrix, and let $\mathbbm{1}\lbrace A \rbrace$ denote the indicator function for an event $A$.
Consider an $n \times m$ data matrix $X$ where each row corresponds to an observation and each column to a variable. We will assume the first $d$ columns are discrete (qualitative or categorical) and the remaining $c=m-d$ columns are continuous (quantitative). We let $X_{ij}$ refer to the element of $X$ in its $i{\text{th}}$ row and $j{\text{th}}$ column. We say $j \in \mathcal{D}$ if $j$ denotes the column index of a discrete variable, i.e.\ $\{1,\ldots,d\}$, and $j \in \mathcal{C}$ if $j$ denotes the column index of a continuous variable, i.e.\ $\{d+1,\ldots,m\}$. Let $b_j$ denote the number of categories for the $j$th qualitative variable. Let $\bm{p}$ denote the vector of qualitative outcome proportions; $\bm{p}$ has length $b = \sum_{j=1}^d b_j$, the total number of categories of all qualitative variables, and ${p}_{j(q)}$ 
is the fraction of observations equal to quality $q$ for the $j$th qualitative variable:

\begin{equation}\label{p}{p}_{j(q)} = (1/n)\sum_{i=1}^n \mathbbm{1}\lbrace X_{ij} = q \rbrace \end{equation} 

We also let $\mu_j$ and $\sigma_j$ be the sample mean and sample standard deviation for each of the continuous variables, $j \in \mathcal{C}$. 

Encodings or transformations of the original column variables are the first step in the FAMD and FAMDAD algorithms. The categorical (i.e.\ discrete) columns are first encoded as a one-hot matrix $Y$. The $j$th categorical variable will utilize $b_j$ columns, one for each outcome level (or category), in the one-hot matrix $Y$. Each of the one-hot columns is an indicator for the $q$th outcome of the $j$th variable, $j(q)$.
\begin{equation}\label{OneHot} Y_{i,j(q)}= \mathbbm{1}\lbrace X_{i,j}=q \rbrace \end{equation}
The binary indicator columns of $Y$ are next scaled based on the category frequencies $\mathbf{p}$, to form the scaled one-hot matrix $Z^D$, defined element-wise as in equation \ref{OneHotRescale}.

\begin{equation}\label{OneHotRescale} 
{Z}^D_{i,j(q)}=Y_{i,j(q)}/p_{j(q)}-1.\end{equation}

 The scaling by $p_{j(q)}$ has the effect of emphasizing observations which take rare outcomes, and the subsequent subtraction by 1 has the effect of de-meaning the columns.
 Pages (\citeyear{pages2014multiple}) notes that an important objective of FAMD is to balance the contribution of continuous and categorical features, allowing them to be compared on the same variance scale.

The quantitative columns are similarly mean-centered and scaled (to have unit sample variance). Thus for each of the quantitative variables, we transform the continuous columns of $X$ into a new matrix $Z^C$, which is defined elementwise as
	\begin{equation}\label{ContTrans} {Z^C}_{ij}:=(X_{ij}-\mu_{j})/\sigma_{j} \quad 
	\forall j \in \mathcal{C} \quad (j \text{ refers to a continuous feature)} \end{equation}
The centered-scaled continuous columns $Z^C$ are concatenated  with the the discrete columns $Z^D$ to form the fully transformed matrix $Z=[Z^D, Z^C]$.

  In the transformed data matrix $Z$ each row now represents an observation vector in $\mathbb{R}^{t}$, where $t=b+c$ is the sum of the total number of categories overall plus the number of continuous columns. In $\mathbb{R}^{t}$ a \textit{weighted} norm denoted $||\cdot||_W$ is used to represent the belief that distances between observations should be adjusted depending on the frequencies associated with the corresponding categories. For $\mathbf{v}, \mathbf{u} \in \mathbb{R}^{t}$, the norm is defined by the quadratic form in equation \ref{eq:WEq}, and the associated distance metric is defined in equation \ref{WMetric}:
  \begin{equation}\label{WMetricW} W=\text{diag}([\mathbf{p}, \mathbf{1}_{c}]) \end{equation}
 \begin{equation}\label{eq:WEq}||\mathbf{v}||^2_W=\mathbf{v}^TW\mathbf{v} \end{equation}
 \begin{equation}\label{WMetric} d^2_W(\mathbf{u},\mathbf{v})=||\mathbf{u}-\mathbf{v}||^2_W=(\mathbf{u}-\mathbf{v})^TW(\mathbf{u}-\mathbf{v}) \hspace{0.5cm} 
 \end{equation}

FAMD \cite{pages2014multiple} defines $W$ as the diagonal matrix denoted in equation \ref{WMetricW}. Thus, each column of $Z$ will not be weighted equally, the one-hot columns are given weights equal to the proportion of observations which take that categorical level and the continuous variables are given weight one. 

The combined implication of scaling  $Z^D$ by inverse frequency in equation \ref{OneHot} and using the norm defined by $W$ in equation \ref{WMetricW} is that the contribution to the weighted distance between two observations by a category they disagree upon is weighted by the sum of inverse frequencies of those categories. Thus if an observation contains a rare category, it will have a large distance from observations which do not possess that rare category. 

In addition to the weighted metric in $\mathbb{R}^{t}$ for the transformed variables (columns) a weighting metric for the observations (rows) is required to implement FAMD. Typically equal weights are applied to each observation using the observation weighting matrix $B=n^{-1} I_n$, and for vectors $\mathbf{w}, \mathbf{z} \in \mathbb{R}^n$ we define
$$ ||\mathbf{w}||^2_B=\mathbf{w}^TB\mathbf{w},  \quad\quad d^2_B(\mathbf{w},\mathbf{z})=||\mathbf{w}-\mathbf{z}||^2_B. 
$$

 The FAMD algorithm rescales data and selects a variable weighting matrix $W$ carefully as to ensure a balance between categorical and continuous variables. However, for the purpose of anomaly detection, the goal is not to balance each dimension equally, but to seek those dimensions which best reveal anomalies. One indication of how useful a continuous variable is for anomaly detection is given by kurtosis. Kurtosis is the fourth standardized (centered and scaled) moment given by 
 $\kappa={\mu^{(4)}}/{\sigma^4}$, where $\mu^{(4)}$ is the fourth central moment and $\sigma^4$ is the square of the second central moment. 
Kurtosis for the univariate normal distribution is 3, whereas distributions with heavier tails have larger kurtosis. Thus, instead of using the original weighting metric $W$ in equation \ref{eq:WEq}, we propose an alternative weighting matrix $W_{\kappa}$ for FAMDAD defined as
 \begin{equation}\label{MKEq}
 W_{\kappa}:=\text{diag}([\mathbf{p},\bm{\kappa}/3]),
 \end{equation}
 in which $\bm{\kappa}$ denotes the length $c$ vector of sample kurtosis from each of the $c$ continuous variables. In practice we cap the sample kurtosis to some limit, $\bar{\kappa}$, set to 10 in all experiments in this paper. Our proposed metric in equation \ref{MKEq} emphasizes the continuous variables with heavy tailed distributions, and discrete variables with rare categories while constructing the wFAMD and FAMDAD embeddings. 
 
 The objective function of FAMD and the FAMDAD embedding is similar to PCA \cite{pages2014multiple}. The output of FAMDAD embedding is an SVD decomposition of $Z=USV^T$ with associated principal coordinates, see Algorithm \ref{FAMDPsuedo} below for psuedocode. The algorithm naively runs in time $O(nt^2)$ for a full SVD decomposition, but can be reduced to $O(ntk)$ to only compute $k$ singular vectors, even in the case when both the first and last singular vectors are required.
\begin{algorithm}
\begin{algorithmic}[1]
\caption{Factorial Analysis of Mixed Data for Anomaly Detection (FAMDAD), an extension of FAMD\cite{pages2014multiple} for anomaly detection.}\label{FAMDPsuedo}
\BState \textbf{Inputs}  $X$ : $n$ $\times$ ($d+c$) \Comment{input data matrix} 
\BState \textbf{Outputs} $F$ : $n$ $\times$ $(b+c)$ \Comment{continuous embedding}
\State $Y \leftarrow \text{one-hot}(X[\cdot,1:d])$ 
\State $\bm{p} \leftarrow \text{column-sum}(Y)/n$
\State $Z^D \leftarrow Y/\bm{p} - 1$
\State $Z^C \leftarrow (X[\cdot,d+1:d+c]-\mu)/\sigma$ \Comment{center-scale}
\State $Z \leftarrow [Z^D,Z^C] $
\State $\bm{\kappa} \leftarrow \text{kurtosis}(X[:,d+1:d+c])$
\State $W \leftarrow \text{diag}([p,\bm{\kappa}/3)$ \Comment{dimension weights}
\State $B \leftarrow n^{-1} \text{diag}(\text{ones}(n))$\Comment{observation weights}
\State $[U,S,V] \leftarrow \text{svd}(B^{1/2}ZW^{1/2})$
\State $F \leftarrow ZW^{1/2}V$
\end{algorithmic}
\end{algorithm}

\section{Anomaly Models and Performance}
We next study the performance of the proposed approach in high dimensional and big mixed data settings under various anomaly models. In each, we do not suppose {\it a priori} which type of anomalies we are searching for. Extreme anomalies and isolated anomalies that are far from all other points will be easy to detect regardless of transformation. Thus we focus on discovering more challening subspace anomalies.. We have found that anomalies tend to live together in subspaces, that is many features will increase together. Although it may sound antithetical that the anomalies are structured, this is commonly found in real datasets \cite{rahmani2017coherence} The PCA machinery of the FAMDAD embedding finds these anomalous subspaces, greatly enhancing anomaly detection performance. We present  two simplified models to demonstrate the effectiveness of the PCA nature behind the FAMDAD approach. The first model analyzes the structured subspace anomalies, whereas the second analyzes when the inliers have low dimensional structure that the anomalies do not adhere to.

\subsection{Larger Subspace Anomalies}
\label{sec:Large Subspace}
It is a common situation that anomalies are larger in a low dimensional subspace \cite{rahmani2017coherence}. We model such a situation as follows:
Assume we are in an $c$ dimensional latent space, where the anomalies are larger on a $s$ dimensional subspace. This subspace is spanned by the columns of an orthonormal matrix $Q$ which is $c$ by $s$
We assume inliers $\mathbf{x}$ are independently and identically distributed (henceforth i.i.d.) isotropically,
\begin{equation}\label{largeInliers}  \mathbf{x} \sim N(0,I_c) \end{equation}

Let $\mathbf{r}$ be a $c$ dimensional vector of i.i.d standard normals. We assume the following distribution for an i.i.d. anomaly $\mathbf{a}$ : 
\begin{equation}\label{largeAnom} \mathbf{a} \stackrel{d}{=} (I_c + \sigma QQ') \mathbf{r} \end{equation}

Where $\stackrel{d}{=}$ denotes equality in distribution. This is equivalent to the following multivariate normal distribution: 
\begin{equation}\label{anomLargeSubspace} \mathbf{a} \sim N(0,(I_c + (\sigma^2 + 2\sigma)QQ')) \end{equation}

Let the anomaly fraction be $0<\delta<1$. That is if 
$\mathbf{x_f}$ is a random datapoint of this model, with probability $1-\delta$ it is an inlier with isotropic distribution given by equation \ref{largeInliers}, and with probability $\delta$ it is an anomaly with distribution given by equation \ref{largeAnom}.
Then the population covariance matrix $C$ of random variable $x_f$ is given by


\begin{equation}\label{largeCovariance} 
\begin{split}C:=\text{Cov}(\mathbf{x_f}) E(\mathbf{x_F},\mathbf{x_F}') =(1-\delta) E(\mathbf{x}\mathbf{x}') + \delta 
E(\mathbf{a}\mathbf{a}') \\ = (1-\delta) I_c + \delta [I_c + \sigma^2 + 2\sigma QQ'] \\= I_c + \delta(\sigma^2 + 2\sigma)QQ'   \end{split}\end{equation}


We define $g =\delta(\sigma^2+2\sigma)$. Then an Eigendecomposition of $C$ is given by 
$C=V\Lambda V'$, where $V = [Q \quad F]$ (block matrix notation, Q is as defined before and $F$ is any orthonormal matrix spanning the orthogonal subspace to $Q$),
and $\Lambda = \text{diag}([m\mathbf{1}_s \quad \mathbf{1}_{c-s}])$, where $m=1+g=1+\delta(\sigma^2+2\sigma)$. That is $\Lambda$ is a diagonal matrix whose first $s$ entries are $m$, and the rest are one. The validity of the eigendecomposition $C=V\Lambda V'=I_c + \delta(\sigma^2 + 2\sigma)QQ$ can be verified by direct multiplication. 

Since we have purely continuous data FAMD reduces to PCA. If the number of observations is sufficiently large than the true eigenvectors $V$ are well approximated by PCA 

Thus the transformed coordinates of a random inlier $\mathbf{x}$, given by $\text{FAMD}(\mathbf{x})$, will have coordinates equal in distribution to $\mathbf{x}$, since $\text{FAMD}(\mathbf{x}) = V'\mathbf{x} \stackrel{d}{=} \mathbf{x}$, which is true since $V'$ is a rotation applied to an isotropically distributed random variable, $\mathbf{x}$.

However anomalies after the FAMDAD embedding will be distributed as $\text{FAMDAD}(\mathbf{a}) = V'a  \stackrel{d}{=} [Q \quad F]  (I_c + \sigma QQ') \mathbf{r}\stackrel{d}{=} [(\sigma+1)r_s \quad r_{c-s}]$
The coordinates are now independent (before they were not), and the first $s$ have variance $(\sigma+1)$, where the rest have variance 1. Thus the first $s$ coordinates of the FAMD embedding are well suited to separate the anomalies, whereas the last $c-s$ contain no signal. Prior to the FAMD embedding, naive distance based algorithms would fail due to the curse of dimensionality, that is all points would have similar distances to the origin.

\subsection{Unstructured Subspace Anomalies}

A second notion of subspace outliers are those that ignore the correlation structure of inliers. To model this scenario we will assume that inliers follow some $c$ dimensional multivariate normal distribution
$$\mathbf{x} \sim \mathcal{N}(0,C) \text{ with } \hspace{0.1cm} C_{j,j}=1 \hspace{0.2cm} \forall j. $$
Here, $\mathbf{x}$ is an inlier generated by a mean zero multivariate normal with symmetric positive semi-definite covariance matrix $C$, which has all diagonal elements equal to 1, and off diagonal elements arbitrary as long as $C$ is a valid covariance matrix (symmetric semi-definite). The condition that $C$ has ones along the diagonal is equivalent to reducing each variable to unit variance, which is done as a preliminary step in FAMD. We let $Q\Lambda Q'=C$ be an eigendecomposition of $C$, and let $\mathbf{r} \sim \mathcal{N}(0,I_c)$ be a vector of $c$ independent standard normal variables, thus $\mathbf{x}\stackrel{d}{=}Q\Lambda^{1/2}\mathbf{r}$. 

We model anomalies $\mathbf{a}$ that are isotropically distributed,
$$ \mathbf{a} \sim \mathcal{N}(0,\sigma^2 I_c)\stackrel{d}{=} \sigma \mathbf{r}. $$

Since this model also is of purely continuous data, the FAMD embedding will be identical to PCA. If we assume that the anomaly fraction is given by $0 < \delta \ll 1 $, then the FAMD embedding for an inlier $x$, denoted by $\text{FAMD}(\mathbf{x})$, is given by 
\begin{equation}\label{Inlier Coordinates} F(\mathbf{x})=V^{'}\mathbf{x}\stackrel{d}{=}V^{'}Q\Lambda^{1/2}\mathbf{r} \stackrel{d}{=}\Lambda^{1/2}\mathbf{r}. 
\end{equation}
 The principal coordinates of a random outlier, denoted by $F(\mathbf{a})$ is given by 
\begin{equation}\label{Anomaly Coordinates}
F(\mathbf{a})=V^{'}\mathbf{a}\stackrel{d}{=}\sigma V^{'}\mathbf{r}\stackrel{d}{=}\sigma \mathbf{r}. 
\end{equation}
Following the FAMD embedding the transformed coordinates of both inliers given by $F(\mathbf{x})$ in equation \ref{Inlier Coordinates} and the coordinates of anomalies given by $F(\mathbf{a})$ in equation \ref{Anomaly Coordinates} have independent normal coordinates. The only difference between them is the magnitude of the variance scaling for each of the components. 

\begin{figure}
\centering
\includegraphics[width=0.5\textwidth]{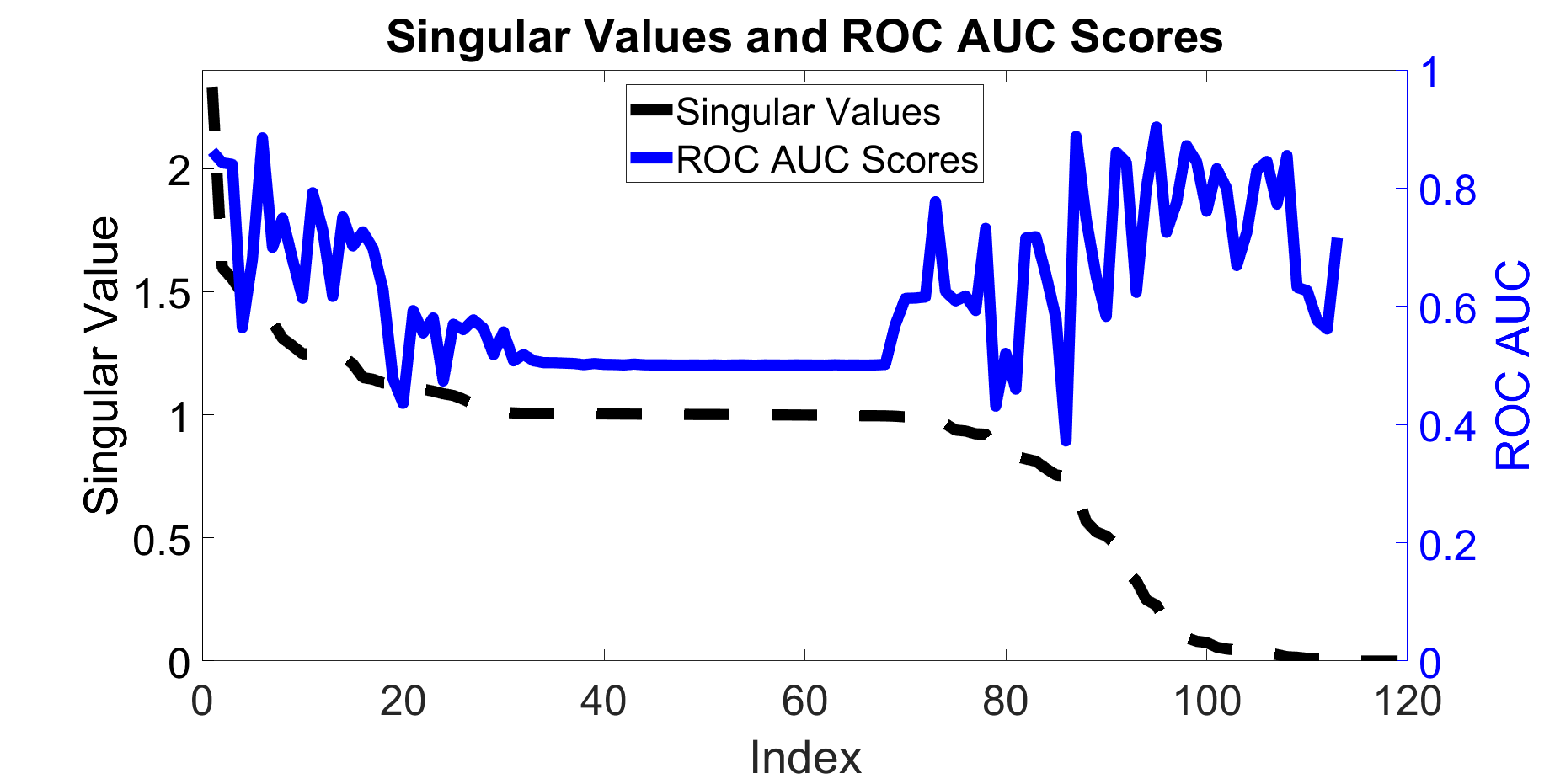}
\caption{\label{fig:kDDSingular} Area Under Curve Receiver Operator Characteristic (AUC ROC) as function of dimension (which single dimension of the FAMD embedding to use) for the KDD dataset. The first 20 and last 40 dimensions give the best AUC score, whereas the middle dimensions give the AUC ROC of a random scoring (0.5). This was a feature found among several datasets, the most useful dimensions of the embedding are the first and the last. The first coordinates are best for separating the large subspace anomalous, whereas the last are required for separating the unstructured subspace anomalies. The singular value spectra is flat due to the presence of many rare categories, the categorical weighting of FAMD will give such features weight nearly one.}
\end{figure}

\begin{figure}
\centering
\includegraphics[width=0.5\textwidth]{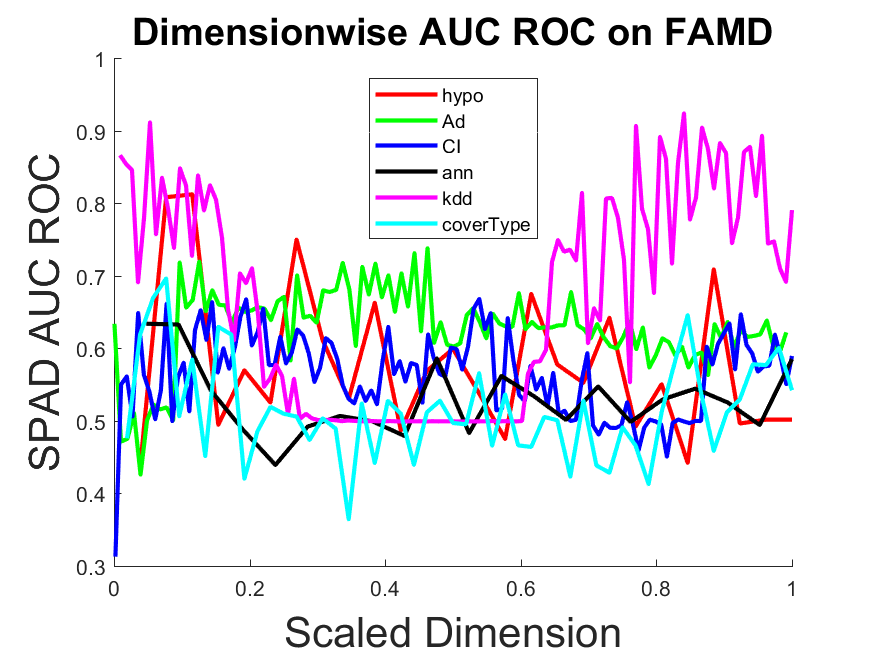}
\caption{\label{SingleW}  SPAD  AUC ROC Scores for the FAMD embedding. The SPAD algorithm is best suited to  generate such a plot due to it's independence assumption. The first and the last coordinates tend to achieve the best AUC ROC, especially for the KDD dataset.}
\end{figure}

Each (random) component of an anomaly, distributed according to equation \ref{Anomaly Coordinates}, is identically distributed with variance $\sigma^2$. The variance of the components of inliers, equation \ref{Inlier Coordinates}, decrease (weakly) monotonically as the diagonal entries of $\Lambda$. We note that $c$=Tr$(C)=\sum_{i=1}^c \Lambda_{i,i}$. Thus the average value of $\Lambda$ is 1. To analyze which components of the final embedding to use, we consider three cases. If $\sigma \gg 1$ then $F(\mathbf{a})$ and $F(\mathbf{x})$ will differ most in the last coordinates, as that is where $F(\mathbf{x})$ has the smallest range and is easiest to differentiate from the larger components generated by normals with variance $\sigma \gg 1$. Similarly if  $\sigma \ll 1$, then $F(\mathbf{x})$ and $F(\mathbf{a})$ will differ most in the first coordinates. 
Interestingly if $\sigma \approx 1$, then $F(\mathbf{x})$ and $F(\mathbf{a})$ will differ most in the first and last coordinates.
This is also motivated from numerical results, such as figure \ref{fig:kDDSingular}. This plot shows that the first dimensions and last dimensions generate the best component-wise Area Under Curve Receiver Operator Characteristic (AUC ROC) scores for the KDD CUP dataset. AUC ROC is a measure of how good a set of scores matches the true anomalies, it can be interpreted as the probability a random outlier is given a higher score than a random inlier. Figure \ref{SingleW} also show how the AUC ROC score varies as a function of ordered dimension number across many real datasets.

\section{Simulations and Experiments} 
We employ a three step approach to detect anomalies. First the mixed data is transformed via FAMD or wFAMD to a purely continuous space. Next, subspace selection is performed. Then the SPAD and Isolation Forest scoring algorithms are applied in the embedded anomaly subspace. There are various modifications of FAMD to consider, such as choice of weighting matrix and which dimensions to use. These choices will be compared, as well as to a baseline of running the algorithms directly on one-hot transformed variables. 
For simplicity we choose to focus on only four variants of FAMD:  FAMD-F (original weighting matrix, first coordinates), FAMD-FL (FAMD with first and last coordinates), wFAMD-F (kurtosis weighted FAMD, first coordinates) and wFAMD-FL (wFAMD with first and last coordinates). For each of these choices of embedding, SPAD or ISO can be run for anomaly scoring. 

We test the embedding properties of FAMD and FAMDAD by applying these embeddings to three synthetic (simulated) datasets. The first demonstrates these method's successful handling of both categorical and continuous outliers. The second shows that FAMD and FAMDAD is able to effectively isolate anomalies whose deviant nature lies only in the interplay between discrete and continuous variables. Such an interaction cannot be captured by any algorithm that considers the discrete and continuous features separately, their correlation must be considered. The third simulation demonstrates FAMD's inherited PCA properties to detect anomalous subspaces in high dimensional data. 

\subsection{First Simulation} 
\begin{figure}
\begin{centering}

    \includegraphics[width=0.22\textwidth, trim=0 0 0 0, clip]{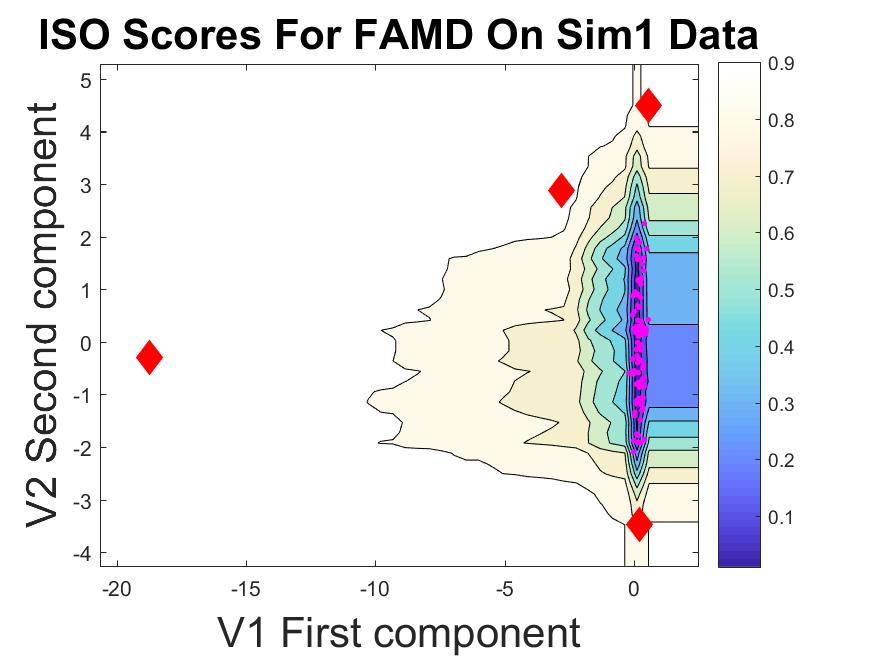} \includegraphics[width=0.22\textwidth, trim=0 0 0 0, clip]{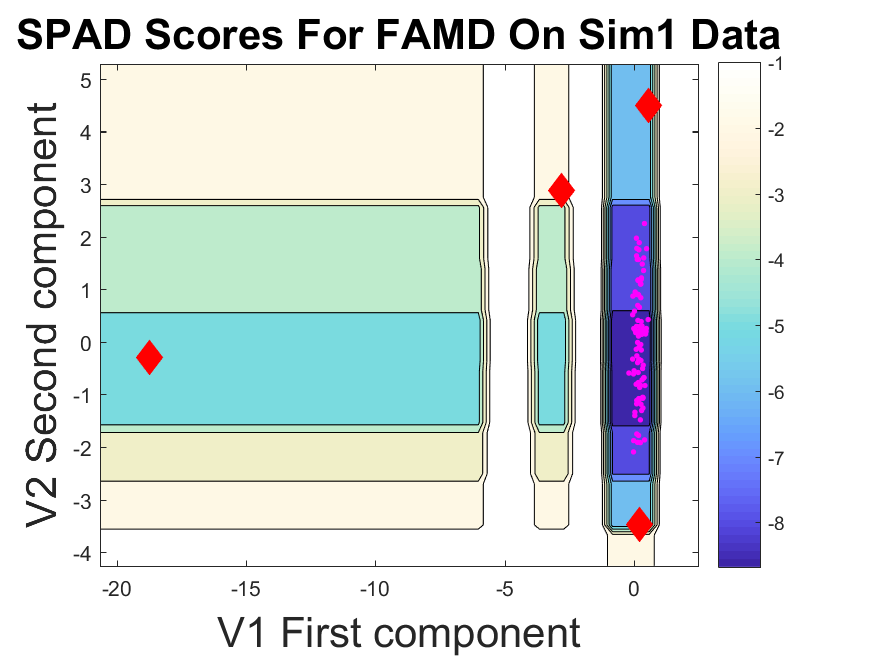} \\
  \includegraphics[width=0.22\textwidth, trim=0 0 0 0, clip]{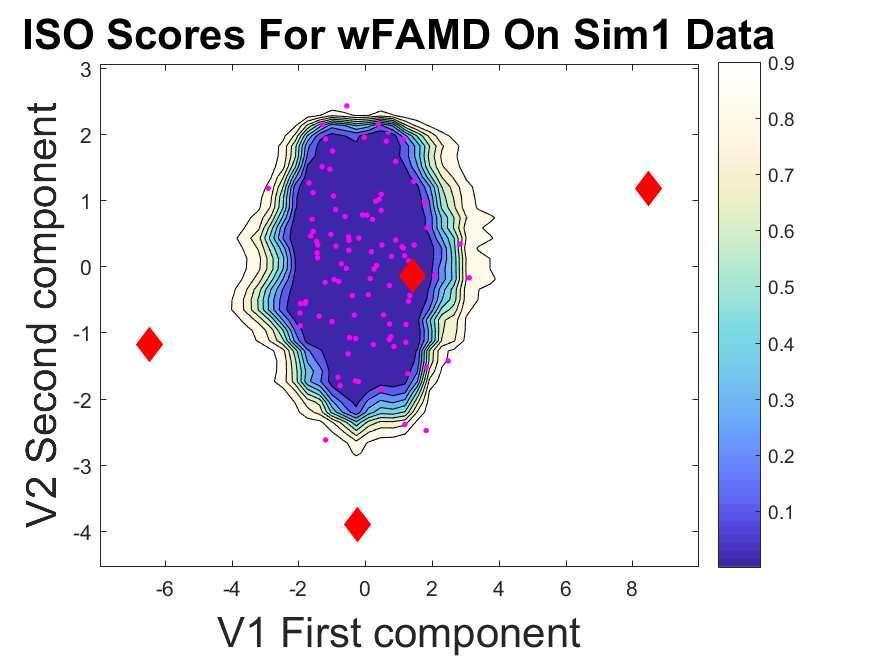}   
  \includegraphics[width=0.22\textwidth, trim=0 0 0 0, clip]{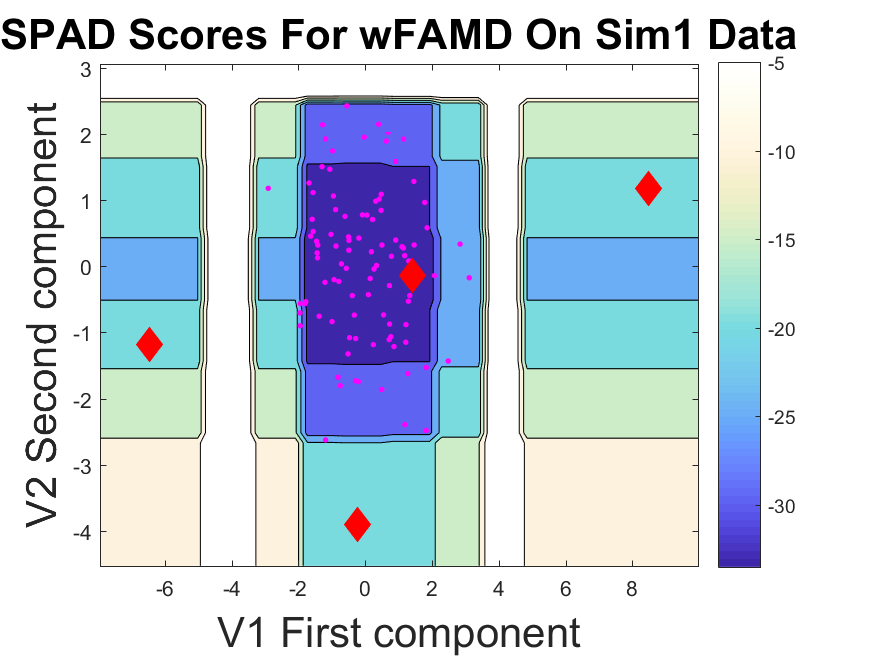}

  \caption{\label{fig:AllResMixed} FAMD and wFAMD embeddings on the Sim1 dataset. The FAMD algorithm has successfully separated the anomalies from the inliers using only the first two principal coordinates. In addition the anomalies can be separated from the inliers with axis-parallel cuts. We see that FAMD is highly sensitive to anomalies, the leading singular vector (the x-axis), points mainly in the direction of the most outstanding outlier. The use of kurtosis in the FAMDAD algorithm gives additional weight to the continuous coordinates due to the continuous outliers. The discrete outlier in the center of inlier cluster is revealed if more than two dimensions are used. SPAD creates blocky score isotherms due to its independence assumption.
  }
  \end{centering}
  \end{figure}

The first simulated dataset has two continuous dimensions both drawn from standard normals and eight binary columns. The data are summarized in table \ref{TableData1}. There are four anomalies, two whose anomalous behavior is that they have abnormally large continuous features, and two that have very rare categorical features.
    \begin{table*}
    \scalebox{1.0}{ 
\begin{tabular}{||c |c |c |c |c |c |c |c |c |c |c ||}
\hline
\textbf{Sim1 Data}     & $X_1$ & $X_2$ & $X_3$ & $X_4$& $X_5$& $X_6$& $X_7$& $X_8$& $X_9$& $X_{10}$                    \\ \hline
i.i.d. Inliers ($100$)     & $N(0,1)$ & $N(0,1)$ & $B(0.5) $ & $B(0.5)$ & $B(0.5)$ & $B(0.5)$ &1 & 1 & 1 & 1\\ \hline
Anomaly 1     & $-1.75$ & $0.34$ & $0$ & $0$ & $0$ & $0$ & \textcolor{red}{0} & \textcolor{red}{0} & \textcolor{red}{0} & \textcolor{red}{0}          \\  \hline
Anomaly 2     & $0.98$ & $0.51$ & $0$ & $1$ & $0$ & $1$ & $1$ & $1$ & $1$ & \textcolor{red}{0}          \\  \hline
Anomaly 3 & \textcolor{red}{5.0}& \textcolor{red}{5.0} & $0$ & $1$ & $1$ & $1$ & $1$ & $1$ & $1$ & $1$                     \\ \hline  
Anomaly 4 & \textcolor{red}{-4.0} & \textcolor{red}{-4.0} & $1$ & $0$ & $1$ & $1$ & $1$ & $1$ & $1$ & $1$                     \\  \hline
\end{tabular}}
  \caption{ {\label{TableData1}} Sim1 Dataset. $B(0.5)$ is a Bernoulli random variable with success probability $0.5$ }
\end{table*}
 %
 %
In order to test the usefulness of the FAMD embedding, we employ two benchmark unsupervised anomaly detectors, SPAD and ISO. SPAD can be employed on mixed data by first discretizing the continuous columns, whereas Isolation Forest requires continuous data. We compare the results of these two scoring algorithms after two embeddings, FAMD and wFAMD, both chosen as two dimensional for visual clarity. From figure \ref{fig:AllResMixed} we see that FAMD successfully separates the anomalies using only two dimensions. We also see that the leading singular vector of FAMD is heavily influenced by anomaly $1$, and to a lesser extent anomaly $2$, without these two anomalies the leading singular vector changes drastically. 
 %
%

%
%
%

\subsection{Second Simulation}
\begin{figure}[b!]
\begin{centering}
  \includegraphics[width=0.22\textwidth, trim=0 0 0 0, clip]{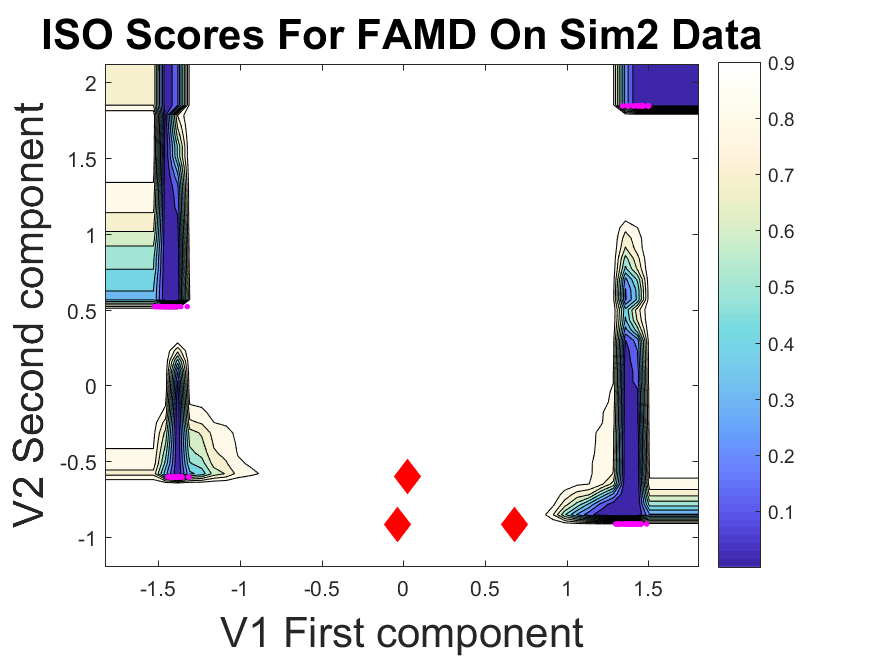}    
  \includegraphics[width=0.22\textwidth, trim=0 0 0 0, clip]{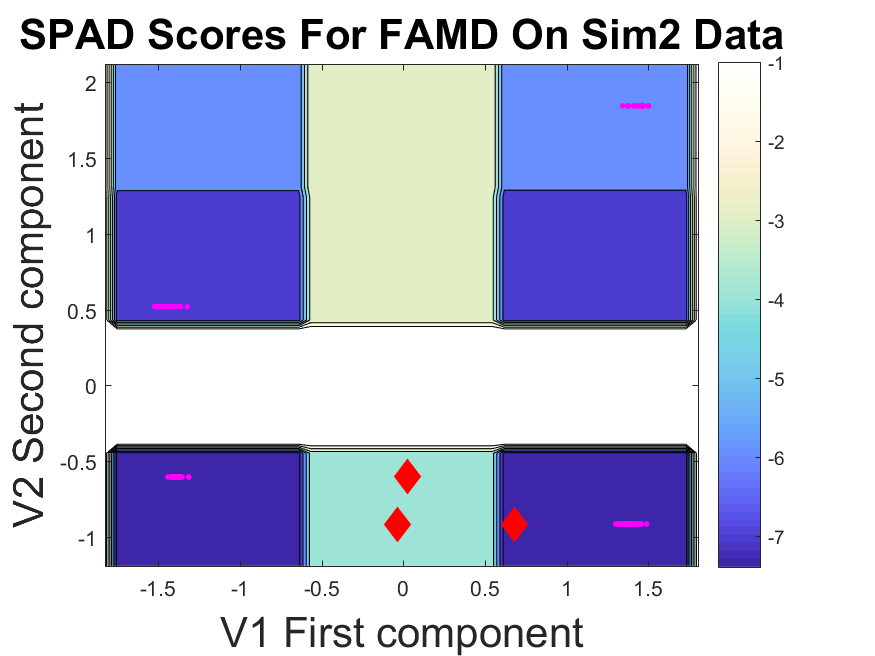} \\
  \includegraphics[width=0.22\textwidth, trim=0 0 0 0, clip]{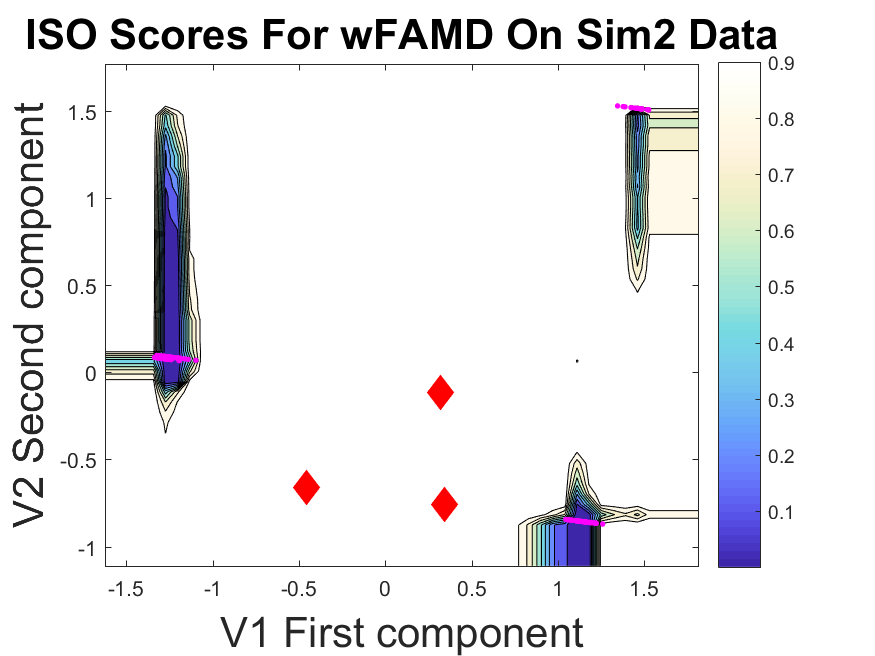}   
  \includegraphics[width=0.22\textwidth, trim=0 0 0 0, clip]{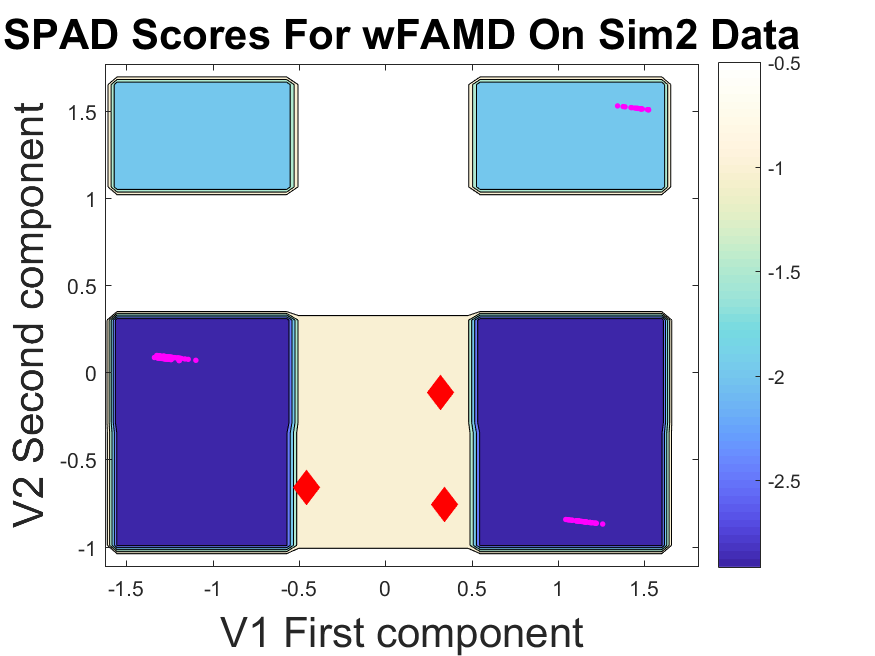}
 \caption{\label{fig:AllResS} FAMD embeddings on the Sim2 dataset. The FAMD algorithm successfully separates the three anomalies using on the first two principal coordinates. The PCA nature of the FAMD embeddings is able to reveal the anomalous structure in single dimensions, while preserving the cluster structure of the inliers.}
 
\end{centering}
\end{figure}

Data used for a second simulation is show in table \ref{TableSSDM}. This dataset demonstrates FAMD's ability to detect subspace anomalies across continuous and discrete dimensions. The anomalies are different from the inliers only in their interplay between their continuous and categorical features. Figure \ref{fig:AllResS} shows the results on this dataset. We see that FAMD and FAMDAD are again able to separate anomalies using only the first two dimensions.

\begin{table}
\scalebox{0.90}{
\begin{tabular}{||c |c |c ||c |c |c |c ||}
\hline
\textbf{Sim2 Data}     & $X_1$    & $X_2$       & $Y_1$  & $Y_2$ & $Y_3$ & $Y_4$          \\ \hline
Cluster 1 (22)     & $N(5,0.1)$       & $1$    & $1$ & $0$ & $0$ & $0$       \\ \hline
Cluster 2 (28)    & $N(5,0.1)$      & $2$  & $0$ & $1$ & $0$ & $0$ \\  \hline 
Cluster 3 (33)    & $N(-5,0.1)$   & $3$ & $0$ & $0$ & $1$ & $0$ \\ \hline
Cluster 4 (17)   & $N(-5,0.1)$   & $4$ & $0$ & $0$ & $0$ & $1$ \\ \hline
Anomaly 1 & \textcolor{red}{0.0} & 3 & $0$ & $0$ & $1$ & $0$ \\ \hline
Anomaly 2 & 5.0 & 3 & $0$ & $0$ & $1$ & $0$\\ \hline
Anomaly 3 & -5.0 & 1 & $1$ & $0$ & $0$ & $0$\\ \hline
\end{tabular}}
    \caption {\label{TableSSDM} Sim2 Dataset. $N(\mu,\sigma):=$Normal with mean $\mu$ and s.d. $\sigma$ \quad Original data of simulated Sim2 is only two dimensional($X_1$ and $X_2$) the one-hot format given by the $(Y_i)$ are used to replace the categorical variable $X_2$}

\end{table}

\subsection{Third Simulation}
\begin{figure}[b!]
\begin{centering}
  \includegraphics[width=0.22\textwidth, trim=0 0 0 0, clip]{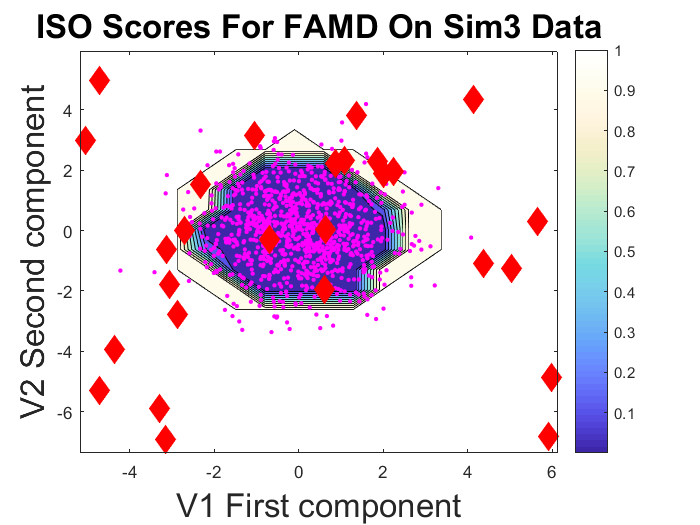}    
  \includegraphics[width=0.22\textwidth, trim=0 0 0 0, clip]{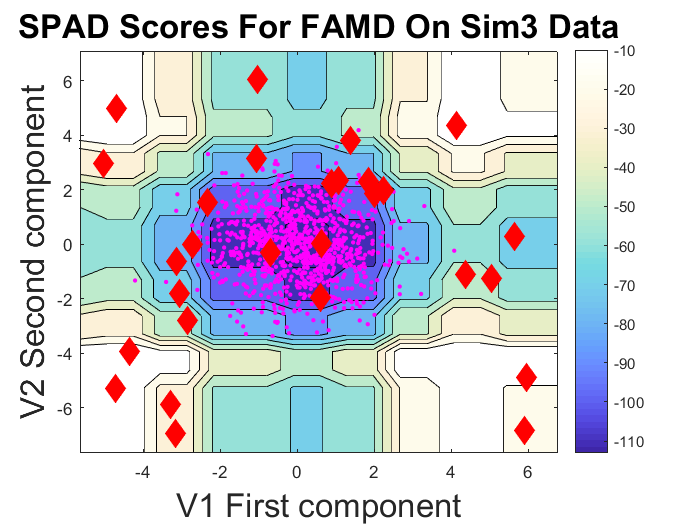} \\
  \includegraphics[width=0.22\textwidth, trim=0 0 0 0, clip]{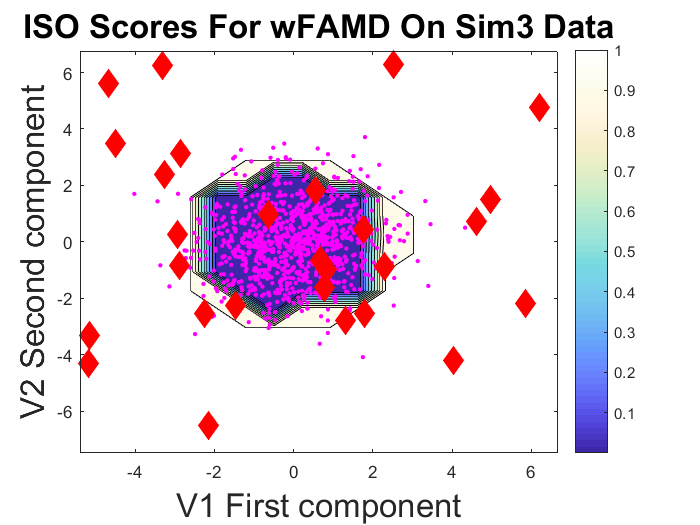}   
  \includegraphics[width=0.22\textwidth, trim=0 0 0 0, clip]{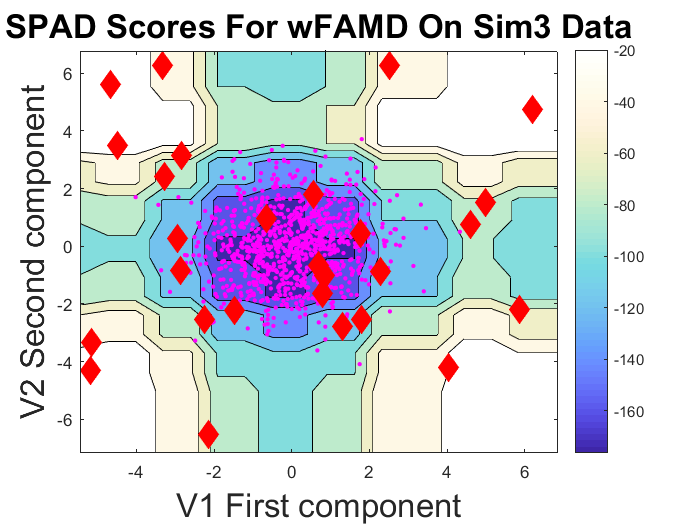}
 \caption{\label{fig:AllResSim3} FAMD embeddings on the Sim3 dataset. Anomalies have larger magnitude coordinates in only a small subspace. Both the FAMD and wFAMD result in good two dimensional representations that enhance anomaly detection performance, these two leading dimensions are made up of linear combinations of the anomalous subspace. We can see that two dimensions is not enough to separate all of the anomalies present, however if more dimensions are used than the anomalies can be nearly perfected separated using these embeddings.}
 
\end{centering}
\end{figure}

We generate data following the larger subspace anomalies analysis as described in section \ref{sec:Large Subspace}. We choose the latent dimension as $c=300$, and a subspace where anomalies will tend to have larger values with dimension $s=10$. We generate a random $s$ dimensional subspace inside of the latent $c$ dimensions by taking the QR decomposition of a random $c$ by $s$ matrix, and using the first $s$ columns of the orthogonal matrix to generate $Q$.
We then generate 1000 i.i.d. inliers with distribution  $\mathbf{x} \sim N(0,I_c)$, and 50 anomalies with distribution $\mathbf{a} \stackrel{d}{=} (I_c + \sigma QQ') \mathbf{r}$, where we choose $\sigma = 3$. Larger magnitudes of $\sigma$ correspond to larger deviations of the anomalous in the $s$ dimensional subspace, while their behavior in the remaining $c-s$ subspace is identical to the inliers regardless of $\sigma$. The results of the two dimensional FAMD and wFAMD embeddings are shown in figure \ref{fig:AllResSim3}. Although the vast majority of the dimensions in the latent space contain no useful information, the first two dimensions of both FAMD and wFAMD begin to reveal the anomalies as having larger deviations from the mean.  

 \subsection{Experiments}

We use the same procedures on several benchmark anomaly detection datasets from the UCI Machine learning repository \cite{asuncion2007uci}. These datasets vary considerably in the number of rows, number of continuous and categorical dimensions, and anomaly percentage as seen in table \ref{table:Datasets Summary}.
\begin{table*}

	  {\scalebox{1.0}{
  \begin{tabular}{||c |c |c |c |c |c |c||}
  \hline
 Dataset & NumRows & Num Dim ($m$) & Num Cont. ($c$) & Num Disc. ($d$) & one-hot Dim ($t$) & anom \%  \\ 
  \hline
 Sim1 & $104$ & $10$ & $2$ & $8$ & 9 & $3.9$\\   \hline
  Sim2 & $103$ & $2$ & $1$ & $1$ & 5 & $2.9$\\   \hline
  Sim3 & $1050$ & $300$ & $300$ & $0$ & 300 & $4.8$\\   \hline
  Covertype & 581012 & 12 & 10 & 2 & 44 & $14.8$\\  \hline
    Annthyroid & 7200 & 21 & 15 & 6 & 44& 7.4\\   \hline
      Hypothyroid & 3163 & 25 & 7 & 18 & 44 & 4.8\\   \hline
      Census & $299285$ & $40$ & $7$ & $33$ & 510 & $6.2$ \\   \hline
 KDD CUP & 50000 & 41 & 34 & 7 & 119 & 2.0\\   \hline
  Ad & 3279 & 1558 & 3 & 1555 & 3114 & 14.1\\   \hline
  \end{tabular}}}
	    \caption{Table summarizing the properties of the datasets used. The non-simulated datasets are sorted by $m=c+d$, the original dimension size. $t$ is the number of dimensions after the one-hot transform. The datasets vary considerably in size, the balance of continuous and discrete features, and anomaly prevalence.}\label{table:Datasets Summary}
  \end{table*}
  %
  \subsection{Dimension Choice}
  To analyze which set of dimensions of the FAMD embeddings to use, we test SPAD on four variants of FAMD, FAMD on first coordinates (FAMD-F), FAMD first and last (FAMD-FL), FAMD using kurtosis-weights on first coordinates (wFAMD-F), and kurtosis weighted first and last (wFAMD-FL). When using the first singular vectors, $k$ refers to using the first $k$ singular vectors. When the first and last singular vectors are used, the first $\ceil*{k/2}$ and last $\floor{k/2}$ singular vectors are used (a total of $k$ singular vectors are used in both cases for easier comparison). 
  
Table \ref{table:Spad Search} shows the result of optimizing $k$ for SPAD, demonstrating that the optimal number of singular vectors to use grows as the original number of features grows. Although the optimal number of dimensions to use is large for the bigger datasets, little accuracy is lost using much fewer dimensions as will be shown in the main results table. 

	  \begin{table*}
	  \begin{center}
	  \scalebox{1.0}{
  \begin{tabular}{||c|c|c|c|c|c||}
  \hline
 Dataset($K$) & NumDim (m) & FAMD-F &  FAMD-FL & wFAMD-F  & wFAMD-FL   \\ 
  \hline
    Sim1  & 10 & 2 & 3 & 3 & 5 \\   \hline
   Sim2  & 2 & 2 & 2 & 1 & 1  \\ \hline
  Sim3 & 300 & 10 & 18 & 10 & 18  \\ \hline
Cover Type & 12 & 9 & 18 & 5 & 9  \\ \hline
Annthyroid & 21 & 2 & 3 & 3  & 3 \\ \hline
Hypo & 25  & 3 & 8 & 1 & 1 \\ \hline
Census & 40 & 140  & 157 & 24  & 48  \\ \hline
KDD & 41  & 112 &  18 & 44 & 27 \\ \hline 
AD & 1558 & 1166 & 1449 & 1558 & 1526 \\ \hline
  \end{tabular}}\newline

  \caption{\label{table:Spad Search}SPAD Search Table. For each dataset embedding pair, SPAD was run and the optimal number of dimensions determined. SPAD works particularly well for determining the optimal number of dimensions because of its independence assumption. NumDim is the original number of features (continuous + discrete) of the original dataset. The FAMD embeddings can be higher dimensional than the original data due to the one-hot embedding of the discrete features. We see that the optimal number of dimensions grows with the dimensionality of the original dataset. Although the optimal number of dimensions for some datasets is fairly large, little accuracy is lost by choosing a smaller dimensional embedding. 
  }
  \end{center}
  \end{table*}

  \begin{center}
\begin{table*}
\scalebox{1.0}{
  \begin{tabular}{||c|c|c|c|c|c|c|c|c||}
  \hline
 & Original  & One-hot &  FAMD-F & FAMD-F & wFAMD-F & wFAMD-F & wFAMD-FL & wFAMD-FL \\
  \hline
  SPAD/ISO & SPAD &  ISO  & SPAD & ISO & SPAD & ISO  & SPAD & ISO  \\ \hline
    Sim1 & 1.00 &  1.00  & 1.00 & 1.00 & 0.98 & 1.00  & 1.00 & 1.00  \\
\hline
    Sim2 & 0.54 &  1.00  & 0.98 & 1.00 & 1.00 & 1.00  & 1.00 & 1.00  \\
\hline
    Sim3 & 0.97 &  0.87  & 1.00 & 1.00 & 1.00 & 1.00  & 0.98 & 0.97  \\
\hline

Cover Type & 0.70&  0.70 & 0.71 & 0.70 & 0.72 & \textbf{0.74}  & 0.70 & 0.70  \\
\hline
Annthyroid & 0.67  & 0.68   & 0.58 & 0.60  & 0.65  &  0.68 &  0.67 & \textbf{0.70} \\ \hline
Hypo & 0.83  & 0.74& 0.76 & 0.72 & 0.82 & \textbf{0.85} & 0.81 & 0.83 \\ \hline
Census &  0.68  & 0.64  & 0.51 & 0.50  & 0.54  &  \textbf{0.73} &  0.70 & \textbf{0.73} \\ \hline
KDD &  0.91 & 0.97 &  0.91 & 0.94 &  0.89 & 0.96 & 0.89 & \textbf{0.97}\\ \hline 
AD & 0.70 & 0.69 & 0.56 & \textbf{0.90} &  0.61 & 0.74 &  0.75 & 0.80\\
\hline
  \end{tabular}}\newline
    \caption{\label{table:Results} Main results table demonstrating the effectiveness of the proposed method using only five dimensions. Two baselines are shown using the entire full dimensional dataset with SPAD and the one-hot embedding with Isolation Forest.The AUC ROC (Area Under Curve of Reciever Operating Characteristic) for each dataset algorithm pair is shown. The best algorithm for each dataset is highlighted in bold. For each dataset this is one of the proposed methods, in addition note that the five dimensional wFAMD-F-ISO did as well or better than plain SPAD or Isolation Forest on the original data in all cases. For larger datasets the accuracy can be furthered increased by using more dimensions, see table \ref{table:Spad Search} for the optimal number of dimensions when using SPAD.}
  \end{table*}
  \end{center}

\subsection{Main Experiment} 
 We have to decide which embeddings to use and which subspaces of those embeddings, as well as which embedding-algorithm pair (SPAD or ISO) to use. We decide to show only three of the four main embeddings (FAMD-F,wFAMD-F,wFAMD-FL) for brevity, and use both SPAD and ISO on those embeddings. The subspace selection is a difficult task in the unsupervised setting. We have explored using the singular value plot like in figure \ref{fig:kDDSingular} to decide. For the FAMD embedding, the average singular value is one. Thus taking coordinates whose singular values correspond to some threshold above and below one is a reasonable choice. Another common unsupervised choice is to choose a set fraction of explained variance (in our case both from the first and last singular vectors). While these ideas were explored we show the results with a simplistic choice of $k=5$ dimensions on the above simulations and datasets. Two baselines are shown for comparison, SPAD run on the original dataset (full dimension), and Isolation forest run on the One-hot encoding of the original dataset (also full dimension). The results are shown in table \ref{table:Results}.

\section{Results and Discussion}
\begin{figure}
\centering
\includegraphics[width=0.47\textwidth]{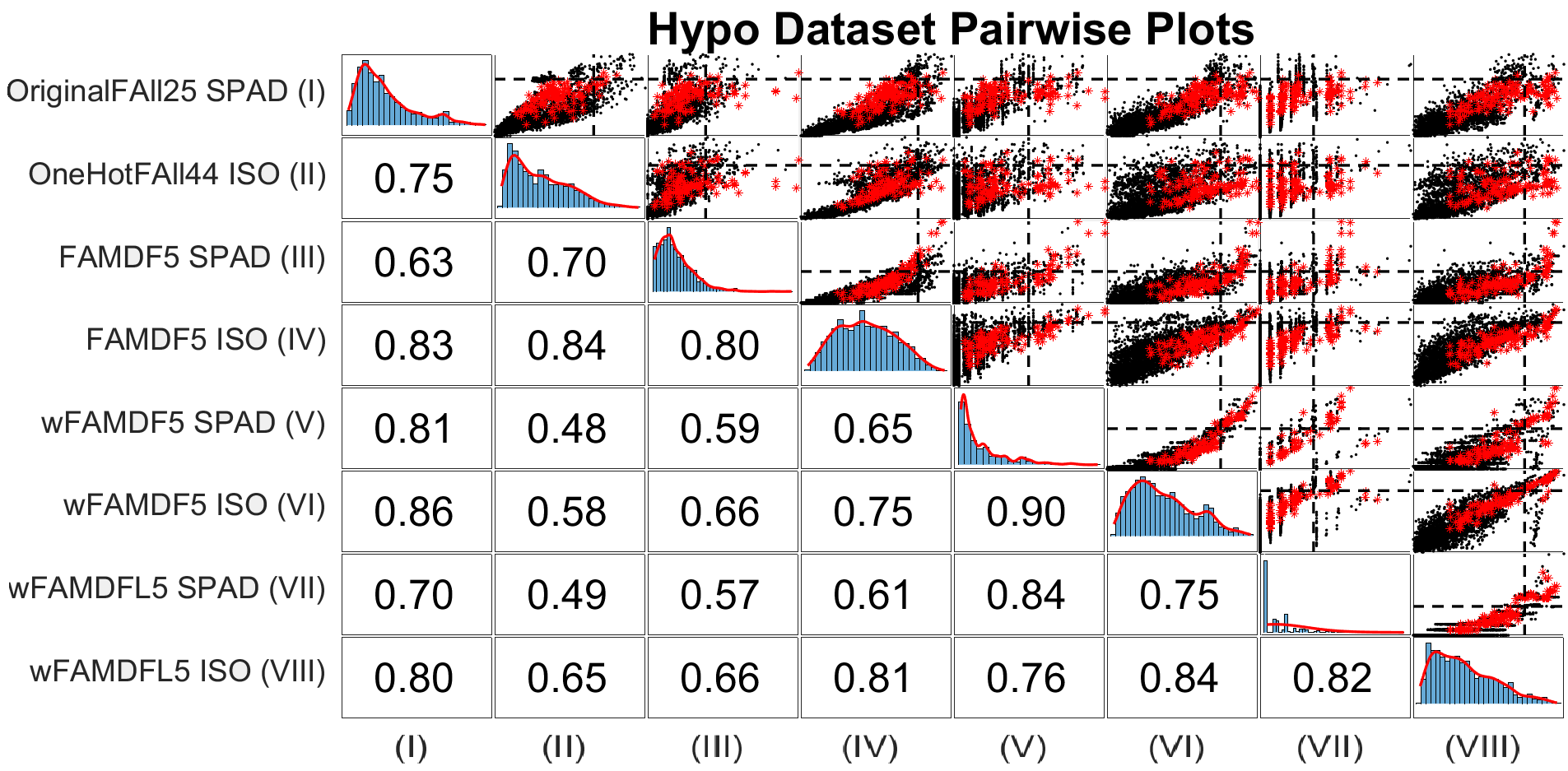}
\caption{\label{fig:HypoPairAll} Pair Plot on Hypo data. In each upper right subplot the x-coordinates are given by the corresponding anomaly scoring algorithm labelled on the top of the plot, and the y-coordinates given by the algorithm on the left. The diagonal subplots show the histogram of scores for that algorithm. The lowest left shows the correlation between the corresponding algorithms. A general trend is stronger correlation with weighted algorithms to each other, and also non-weighted with other non-weighted algorithms.}
\end{figure}

\begin{figure}
\centering
\includegraphics[width=0.5\textwidth]{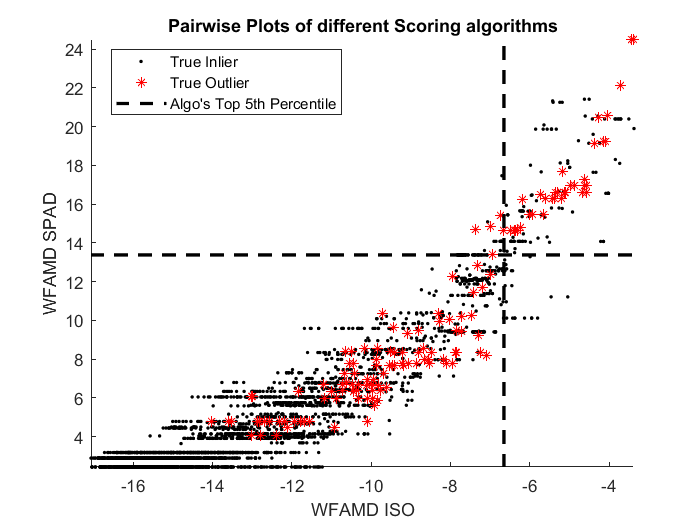}
\caption{\label{fig:HypoPair} Pair Plot for ISO on W5 and SPAD on W5. Red points correspond to ground truth anomalies. There is a non-linear relationship in the graph since SPAD scores have heavier tails (anomalies are ranked much larger than inliers) as compared to Isolation Forest Scores. The general monotonic relationship show the two algorithms mainly agree, and perform well since most of the red points are located in the top right.}
\end{figure}

\begin{figure}
\centering
\includegraphics[width=0.5\textwidth]{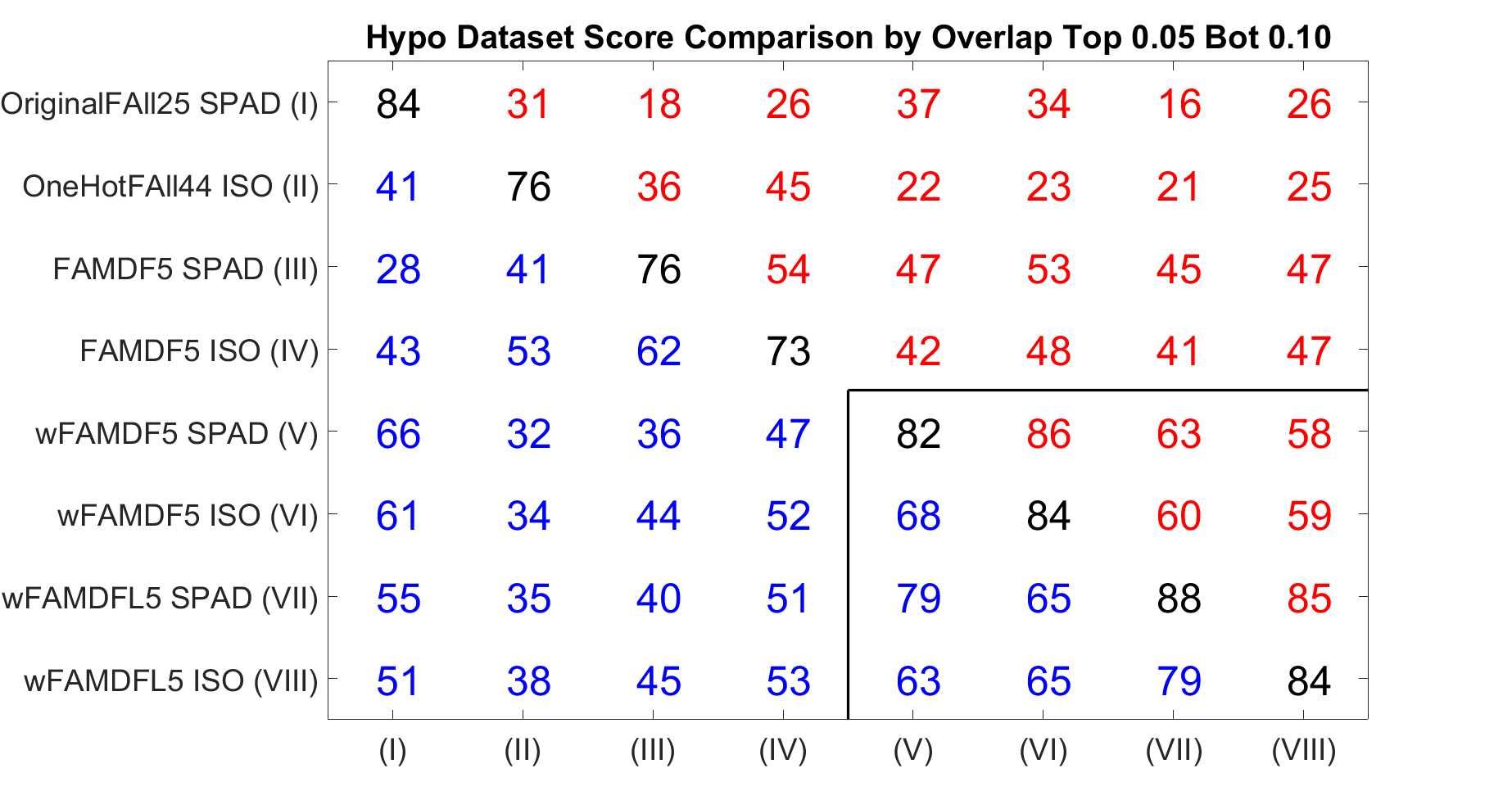}
\caption{\label{fig:HypoPairAllAlt} An alternative pair plot on Hypo data. In each off-diagonal subplot the overlap is defined as the fraction of common anomalies among the top 5 percent (top right) and top 10 percent (bottom left) of each algorithm. The diagonal subplots are the AUC ROC scores for that algorithm. We see a clustering (the bottom right square) of high overlap due to the use of the kurtosis weighting. The larger values on the lower half of the diagonal show that the kurtosis weighting is an effective unsupervised feature selection method for anomaly detection.}
\end{figure}

  For each dataset, a proposed method is shown to be superior to the default methods of the original features after one-hot transformations as shown in table \ref{table:Results}. In some cases, such as the AD and Cover Type datasets, this difference is drastic.
To get a sense of the comparison between different algorithms, we consider analyzing the top scoring observations as ranked by the different anomalies algorithms. Note that other measures like correlation of the rankings are not particularly meaningful here, as one is usually only interested in how well the algorithms rank the top scores, which correspond to the higher anomalous observations. For instance each algorithm can rank true inliers somewhat randomly among the bottom ranks, creating large differences in correlation despite finding the same top anomalies.
Pairwise plots are shown in figure \ref{fig:HypoPairAll}. These are the anomaly scores, where the $x$ coordinates are the scores from one algorithm, and $y$ coordinates from another. True anomalies are plotted in red. The major patterns one can see is a linear relationship when an ISO is compared to an ISO (bottom right) or SPAD to SPAD (top left), but slightly nonlinear when SPAD is compared to ISO. This is due to inherent differences of distributions in scores (seen along diagonal). A single comparison plot is shown in figure \ref{fig:HypoPair} which is the pair plot for ISO on wFAMD-F and SPAD on wFAMD-F. An alternative visualization of overlap measure is shown in figure \ref{fig:HypoPairAllAlt}. We see that weighting by kurtosis creates a large change in the top rankings of scores, each variant using the kurtosis weighting are similar to each other but different than the rest while having higher AUC ROC scores.

\section{Conclusion}
Most anomaly detection algorithms can only handle purely continuous or purely discrete data. The proposed three step FAMDAD approach detects anomalies in a high dimensional unsupervised mixed data setting. The first step embeds the data into a continuous space. Then an anomaly subspace is selected. Finally, a classical anomaly detection method is applied in the lower dimensional embedded space. Our embedding tends to align the data along the major axis, increasing the validity the independence assumption in SPAD, and allowing isolation forest to better isolate data anomalies with axis-parallel cuts, increasing the detection performance of both scoring strategies. Our approach is shown effective on several high dimensional benchmark datasets, outperforming SPAD or Isolation Forest on the one-hot matrix while using a vastly smaller number of dimensions. The choice to weight features by their kurtosis is an effective unsupervised approach to emphasize important features. By using both the first and the last coordinates, the FAMDAD embedding is able capture both anomalies that have large coordinates in low dimensional subspaces, and subspace anomalies that do not adhere to the typical covariance structure.
Open questions include how to automate subspace selection and how to combine the scoring from multiple scoring strategies. 


\section*{Acknowledgements} 

Financial support is gratefully acknowledged from the Cornell University Institute of Biotechnology, the New York State Division of Science, Technology and Innovation (NYSTAR), a Xerox PARC Faculty Research Award, National Science Foundation Awards 1455172, 1934985, 1940124, and 1940276, USAID, and Cornell University Atkinson’s Center for a Sustainable Future.

\bibliographystyle{ACM-Reference-Format}
\bibliography{bibliography}

\end{document}